\documentclass[10pt,twocolumn,letterpaper]{article}

\usepackage{cvpr} 
\usepackage{multirow}

\usepackage[dvipsnames]{xcolor}

\definecolor{cvprblue}{rgb}{0.21,0.49,0.74}
\usepackage[pagebackref,breaklinks,colorlinks,citecolor=cvprblue]{hyperref}

\def\paperID{*****} 
\def\confName{CVPR}
\def\confYear{2024}

\title{Fewer Tokens and Fewer Videos: Extending Video Understanding \\ Abilities in Large Vision-Language Models}

\author{Shimin Chen\thanks{These authors contributed equally to this work.}, \quad Yitian Yuan\footnotemark[1], \quad Shaoxiang Chen,\quad Zequn Jie\thanks{Corresponding Author.},\quad Lin Ma\\
Meituan Inc. \\
{\tt \small chenshimin@zju.edu.cn \quad  yuanyitian@foxmail.com \quad sxchen13@fudan.edu.cn }\\
{\tt \small zequn.nus@gmail.com \quad forest.linma@gmail.com }
}

\begin{document}

\maketitle

\begin{abstract}
Amidst the advancements in image-based Large Vision-Language Models (image-LVLM), the transition to video-based models (video-LVLM) is hindered by the limited availability of quality video data. This paper addresses the challenge by leveraging the visual commonalities between images and videos to efficiently evolve image-LVLMs into video-LVLMs. We present a cost-effective video-LVLM that enhances model architecture, introduces innovative training strategies, and identifies the most effective types of video instruction data. Our innovative weighted token sampler significantly compresses the visual token numbers of each video frame, effectively cutting computational expenses. We also find that judiciously using just 10\% of the video data, compared to prior video-LVLMs, yields impressive results during various training phases. Moreover, we delve into the influence of video instruction data in limited-resource settings, highlighting the significance of incorporating video training data that emphasizes temporal understanding to enhance model performance. The resulting Fewer Tokens and Fewer Videos LVLM (FTFV-LVLM) exhibits exceptional performance across video and image benchmarks, validating our model's design and training approaches.

\end{abstract}

\section{Introduction}
\label{sec:intro}


Large Vision-Language Model (LVLM) have witnessed remarkable advancements in the field of image understanding. Recent studies~\cite{liu2023visual,liu2023improved,instructblip,chen2023minigpt,ye2023mplug,zhu2023minigpt} utilize large language models (LLMs) to interpret images as text-like tokens, excelling in tasks such as image captioning and visual question answering. The success of image-based LVLMs is fueled by the rich diversity and availability of image data. However, adapting these models to video data poses challenges due to the temporal dynamic nature of videos and the hurdles in amassing extensive video datasets.


\begin{figure}
  \centering
   \includegraphics[width=1.0\columnwidth]{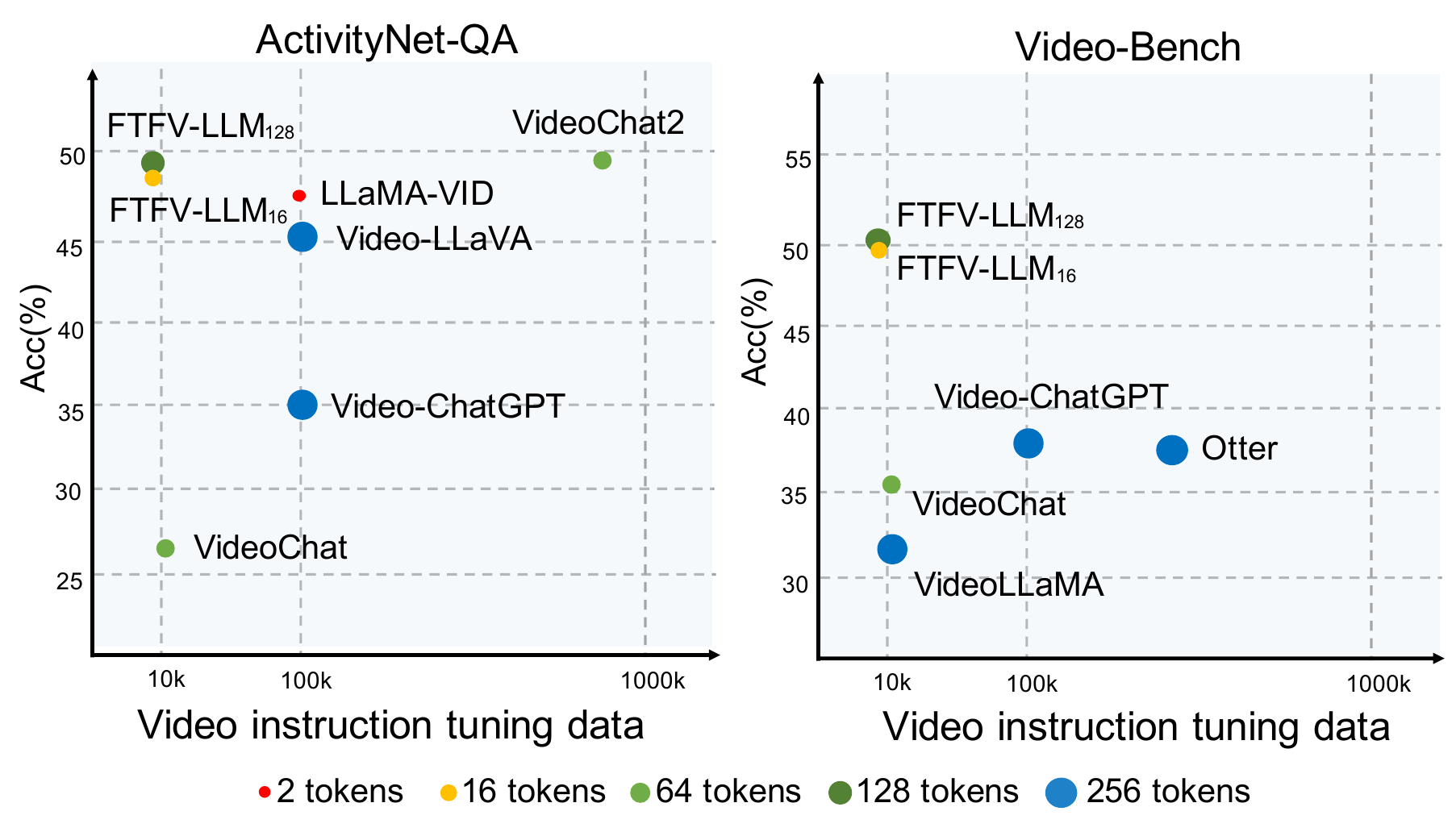}
   \caption{The comparison of our proposed FTFV-LLM to existing video-LVLMs on ActivityNet-QA~\cite{caba2015activitynet} and Video-Bench~\cite{ning2023video}. The horizontal axis indicates the quantity of QA pairs utilized for video instruction tuning in each model, while the vertical axis shows the accuracy achieved on the two benchmarks. Different colors are used to distinguish the number of tokens representing each video frame in these models. Our FTFV-LLM, which utilizes fewer video tokens and fewer video training data, achieves leading results over previous video-LVLMs. Meanwhile, our FTFV-LLM with different token numbers (16 \textit{vs.} 128) also perform similarly.} 
   \label{fig:motivation}
\end{figure}


Recent video-LVLMs~\cite{zhang2023video,li2023videochat,li2023llama,lin2023video,luo2023valley,maaz2023video,liu2023one,song2023moviechat} have emerged by enhancing image-LVLMs with temporal dynamics and long-duration video processing capabilities. Despite efforts to curate video datasets, the volume of video data lags behind image data. Given that videos are sequences of image frames, we investigate the potential of a robust image-LVLM in the video realm. Moreover, considering the temporal and contextual strengths of LLMs, and the excellent visual semantic perceiving abilities of image-LVLMs, we explore cost-effective ways to extend image-LVLMs to the video domain.

To this end, we begin with a foundational image-LVLM and explore methodologies to endow it with video comprehension capabilities. Specifically, the image-LVLM is composed of a Vision Transformer (ViT) visual encoder, a vision-language adapter, and a decoder LLM. A classical three-stage training process is adopted to tune the model, which consists of a multimodal aligning stage, a multimodal pretraining stage, and an instruction tuning stage.  This enables us to develop a proficient image-LVLM.


We then evolve our image-LVLM into a video-LVLM by focusing on model architecture and training strategy. Architecturally, we approach a video as a series of frame images, encoding each frame with the same visual encoder used for images. In this work, we refrain from explicitly modeling temporal relationships within videos, instead relying on the LLM's inherent temporal and contextual processing abilities by sequentially feeding video tokens into the LLM. Additionally, to address the higher computational cost and redundancy across video frames, we introduce a \textit{weighted token sampler} module. This module selectively retains tokens with higher visual attention responses, 
and therefore makes much \textbf{fewer video tokens} input to the LLM. This strategy has proven to be efficient in cutting computational demands and preserving model performance.

Regarding the training strategy, we hypothesize that a powerful image-LVLM may not require an extensive amount of video training data to perform well on certain video understanding tasks, especially those that do not demand temporal reasoning but rather rely on basic visual recognition and perception. Consequently, we test the image-LVLM directly on various video benchmarks and find that it indeed achieves leading results on benchmarks that are less focused on temporal reasoning and action recognition.
We further explore various training strategies by integrating a modest amount of video data (approximately 10\% of what was previously used) with image data across different training phases. It is surprising to find that our model attains high performance on several video benchmarks using significantly \textbf{fewer videos} than earlier approaches, as shown in Figure~\ref{fig:motivation}. We also find that it is not necessary to include video data in the early stages (align and pretrain) of model training, only in the instruction tuning stage, or adding an additional video instruction tuning stage after the three-stage training of image-LVLM can also achieve good results. These findings confirm that an image-LVLM with strong visual understanding can efficiently transition to a video-LVLM at a low cost.



Based on the above findings, we further delve into identifying the most impactful types of video instruction data for enhancing an image-LVLM to a video-LVLM with limited videos. Our research reveals that extensive diversity in video data type is not quite essential for this advancement. This stems from the pre-existing visual knowledge acquired from image training, which often overlaps with video content. Crucially, our studies underscore the value of including video data that focuses on temporal reasoning -- a feature absent in still images. Moreover, to preserve the model's ability to generalize, it's important to incorporate a variety of scenarios in videos, alongside diverse instructional content and formats in question-answering tasks.

The primary contributions of this paper are as follows:


\begin{itemize}
\item We introduce an enhanced large video-language model FTFV-LLM based on a foundational image-LVLM, featuring a novel weighted token sampler module. This module significantly condenses the visual tokens of each video frame, reducing computational requirements of the model while maintaining its efficacy.
\item We propose optimized training strategies for video-LVLM that require minimal video data. Our findings reveal that our model, even with significantly fewer training videos than previous methods, delivers superior results in various video and image benchmarks. This highlights the intrinsic visual perception capabilities of image-LVLMs for video understanding.
\item We examine the impact of different video instruction data types when video resources are limited. Contrary to traditional beliefs, we find that video data centered on temporal reasoning, coupled with comprehensive instructional content, offers more value than a larger scope of video data.
\end{itemize}

\section{Related work}
\label{sec:related work}

\subsection{Large Language Models}
The natural language processing community has been revolutionized by the emergence of Large Language Models (LLMs) in recent years.These models, such as GPT\cite{radford2018improving}, OPT\cite{zhang2022opt}, LLaMA\cite{touvron2023llama}, and MOSS\cite{sun2023moss}, have demonstrated extraordinary abilities in language generation and understanding complex task. To maximize the proficiency of LLMs, researchers have focused on instruction tuning, a key strategy that aligns the model with user instructions and improves output quality. Examples of this approach include InstructGPT\cite{ouyang2022training}. Open-source LLMs like LLaMA\cite{touvron2023llama}, have further inspired works such as Alpaca\cite{alpaca}, Vicuna\cite{chiang2023vicuna}, and ChatGLM\cite{du2022glm}. Their large model capacity allows them to handle various complex tasks and align with human language instructions and preferences.



\subsection{Large Vision-Language Models}

Extending LLMs to multi-modal, especially involving vision inputs such as images and videos, have also attracted a lot of attention. We refer to such models as Large Vision-Language Models~(LVLM), and further split them to image-LVLMs and video-LVLMs according to their inputs.

\noindent \textbf{Image-LVLMs}: 
Pioneering works such as Flamingo\cite{alayrac2022flamingo} have demonstrated the ability to handle sequences of interleaved visual and textual data, allowing for seamless ingestion of images or videos as inputs. This was achieved through training with image-text interleaved datasets. Subsequently, BLIP-2\cite{li2023blip2} proposed an efficient pre-training strategy that bootstraps language-image pre-training with frozen image encoders and LLMs. This strategy has shown promising results in enhancing the performance of pre-trained models in the context of image-LVLMs. Furthermore, LLaVA\cite{liu2023llava} made an attempt to extend instruction-tuning to multimodal tasks. InstructBLIP\cite{instructblip} and MiniGPT-4\cite{zhu2023minigpt} have further enhanced pre-trained models by constructing high-quality instruction pairs based on the principles introduced in BLIP-2. 
In summary,, these works have contributed to the progression of image-LVLMs, introducing novel approaches and techniques to handle interleaved visual and textual data, efficient pre-training strategies, instruction-tuning in multimodal tasks, and the construction of high-quality instruction pairs. Following the previous work, we have developed a high-performance basic image-LVLM based on LLaVA-Next\cite{liu2024llavanext}.

\noindent \textbf{Video-LVLMs}:
Video-LVLMs commonly employ visual encoders to extract video features, temporal modeling modules to encode temporal information, and large language models for generating results. Specifically, Video-ChatGPT has explored the use of CLIP for video embedding extraction, while BLIP-2 has been utilized in VideoChat, LLaMA-VID, among others. Additionally, researchers have focused on temporal modeling, as seen in the case of Video-LLaMA that employs Qformer for temporal modeling. VideoChat integrates the temporal modeling strategies from traditional video classification into multimodal large-scale models. However, processing long video sequences remains a challenge for these models due to the substantial number of tokens required. To tackle this issue, LLaMA-VID efficiently encodes each frame with just 2 tokens, enabling the understanding of long videos in existing language models. Moreover, in terms of training strategies, joint training techniques, as employed in LLaMA-VID and Video-LLaMA, have proven successful in empowering large language models to comprehend videos. Additionally, some studies have focused on pretraining features. Expanding large language models to additional visual modalities necessitates pre-alignment, as demonstrated by LLaMA-Adapter and ImageBind-LLM. These models highlight the advantages of a unified feature space in enhancing the multimodal reasoning capabilities of large language models.

Although these approaches have achieved high performance through model innovation and the construction of large-scale datasets, the fundamental questions of how much training data a good video model truly requires and what kind of data is necessary have remained unexplored. In this paper, we aim to extend the power of an image-based LLM to a video-based LLM with reduced computational cost and less video data.

\begin{figure*}[!t]
  \centering
   \includegraphics[width=1.0\linewidth]{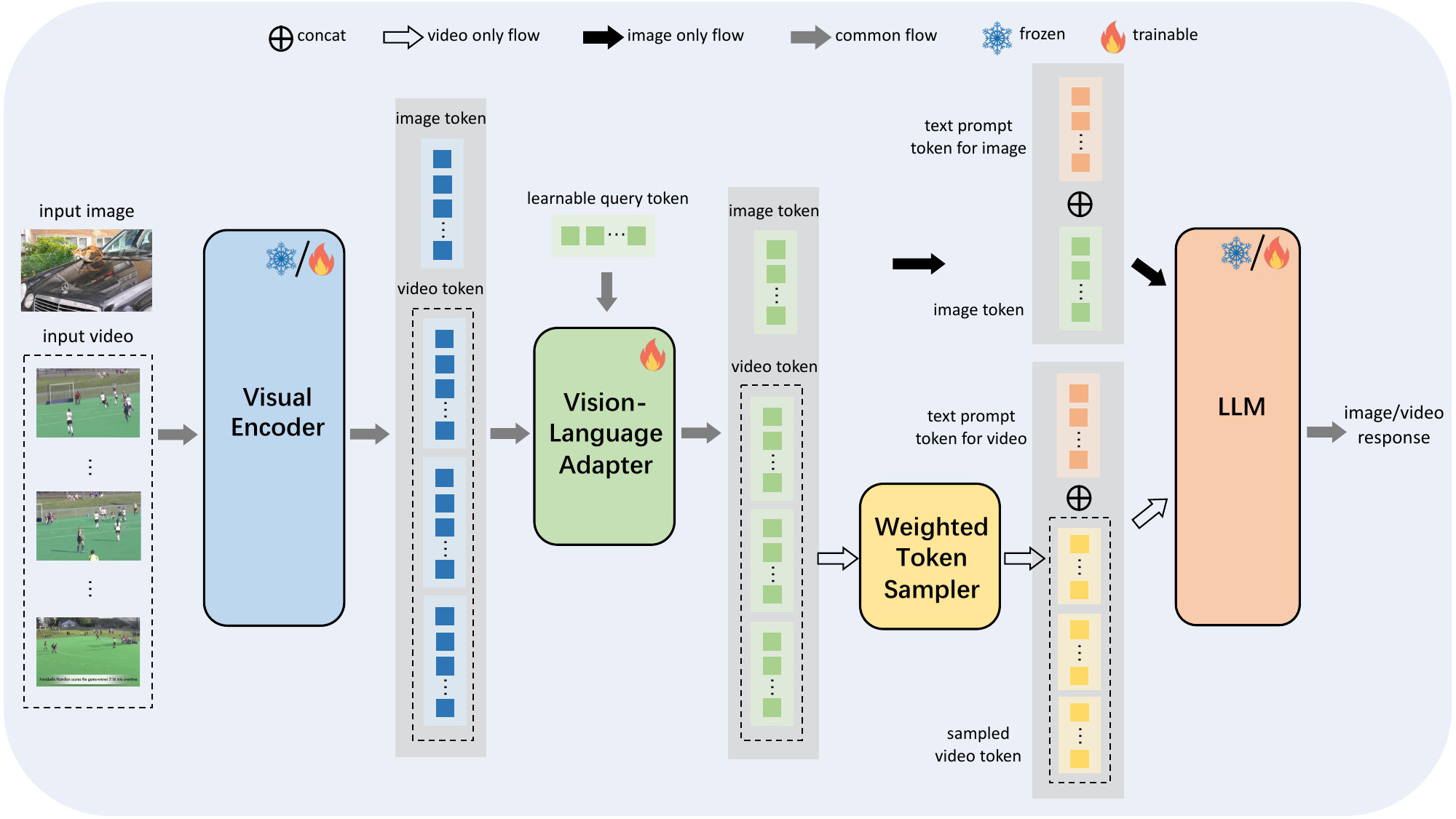}
   \caption{An overview of our proposed FTFV-LLM model. The FTFV-LLM extends from a basic image-LVLM architecture, with a visual encoder firstly encodes each video frame, and then a vision-language adapter modulates the video tokens to align with the LLM feature space. Besides, we propose a novel weighted token sampler module, which can largely compress the token numbers of each video frame, and thus it is beneficial to save the calculation cost of the model when processing multiple video frames. Finally, the compressed video tokens as well as the text prompt tokens are feed to the LLM, thus getting the response output. During our training process, we finetune the model using a combination of video and image data, with comprehensive details available in the main paper.} 
   \label{fig:framework}
\end{figure*}

\section{FTFV-LLM}

In this section, we will introduce our proposed FTFV-LLM model, with first presenting our basic image-LVLM architecture and its corresponding three-stage training pipelines. Then, we will further extend the image-LVLM to a video-LVLM by adding a weighted token sampler module and exploring different video-incorporated training strategies. The overall architecture of our FTFV-LLM is shown in Figure~\ref{fig:framework}.

\subsection{The Basic Image-LVLM}

\subsubsection{Model Architecture}

The framework of our basic image-LVLM follows most of the previous works~\cite{instructblip,zhu2023minigpt,liu2023visual,bai2023qwen} and mainly contains a ViT-based~\cite{dosovitskiy2020image} visual encoder, a vision-language adapter, and a LLM text decoder.

Given an image $I \in \mathbb{R}^{H \times W \times 3}$, the visual encoder is first employed to produce the image token features $\mathbf{I} \in \mathbb{R}^{M \times D}$, where $M =H/p \times W/p$ and $D$ indicate the number of image tokens (patches) and token dimensions, respectively. Our visual encoder takes the Openclip's ViT-bigG~\cite{ilharco_gabriel_2021_5143773} architecture and the patch size $p$ is typically set to 14.

To bridge the gap between the visual encoder and the LLM, and also alleviate the efficiency issues arising from long image token sequences, we leverage a cross-attention module as the vision-language adapter. The module first projects the image token features $\mathbf{I}$ to a new embedding space of dimension $C$ with one linear projection layer, yielding $\mathbf{\Tilde{I}} \in \mathbb{R}^{M \times C}$. Then, it maintains $N=256$ learnable vectors $\mathbf{Q} = [\mathbf{q}^1, \cdots, \mathbf{q}^i, \cdots, \mathbf{q}^N ] \in \mathbb{R}^{N \times C}$ and uses it as queries, $\mathbf{\Tilde{I}}$ as keys/values for cross-attention operations, and thus gets the compressed image token features $\mathbf{X}_I \in \mathbb{R}^{N \times C}$. Meanwhile, 2D absolute positional encodings are also introduced into the cross-attention’s query-key pairs to reduce the potential loss of positional details during compression. 

Finally, the compressed image tokens, together with the text tokens of input question are feed into the LLM decoder, making it output the answer responses. Specifically, we utilize the Vicuna-7B~\cite{vicuna2023} as our decoder, which is
built upon LLaMA~\cite{touvron2023llama} and can perform a wide range of complex linguistic tasks.

\subsubsection{Training Pipeline}\label{capter: 3.1.2}

To achieve an effective image-LVLM, we adopt a three-stage training pipeline, which composes of a \textit{multimodal aligning stage}, a \textit{multimodal pretraining stage}, and an \textit{ instruction tuning stage}.

In the first multimodal aligning stage, we freeze both the visual encoder and LLM, and only tune the vision-language adapter with a large collection of aligned image-text pairs from the LAION~\cite{laion-400m} dataset. This stage is to ensure the visual features are well aligned with the language space.

The second multimodal pretraining stage is designed to 
make the model acquire vision-language knowledge in-depth, from a series of datasets that need more detail visual understanding and more complex reasoning. Therefore, we combine grounding data from GRIT~\cite{Kosmos2} and Visual Genome~\cite{krishna2017visual},  VQA data from VQAv2~\cite{goyal2017making} and GQA~\cite{hudson2019gqa}, and caption data from LAION~\cite{laion-400m} and ShareGPT4V~\cite{chen2023sharegpt4v} to train our model.  In this stage, the last 3 layers of ViT, the vision-language adapter and the LLM are unfrozen and jointly optimized.

In the third instruction tuning stage, we focus on enhance the model's ability in complex instruction following, which is proven to be crucial for LLMs. Specifically, we leverage the recent collected multimodal instruction-following datasets ShareGPT4V~\cite{chen2023sharegpt4v}, M3IT~\cite{li2023m3it} to train our model, and the modules being tuned are consistent with the second stage. The datasets used in the overall training process of image-LVLM are summarized in Table \ref{tab:image_training_data}.

\subsection{The Extended Video-LVLM}

In order to extend the above basic image-LVLM to a video-LVLM, we first evenly sample $T$ frames from the input video, and then feed each frame to the visual encoder independently, getting the corresponding video token sequences $\mathbf{V} = [\mathbf{v}^1, \cdots,  \mathbf{v}^t, \cdots, \mathbf{v}^T] \in \mathbb{R}^{T \times M \times D}$. In order to record the temporal relationship of video frames, we add a group of learnable temporal encoding vectors $\mathbf{E} \in \mathbb{R}^{T \times 1 \times D}$ to the video token features, and thus obtain the temporally-fused video tokens $\mathbf{\hat{V}}=[\mathbf{\hat{v}}^1, \cdots,  \mathbf{\hat{v}}^t, \cdots, \mathbf{\hat{v}}^T] \in \mathbb{R}^{T \times M \times D}$. The vision-language adapter then transforms and compresses each $\mathbf{\hat{v}}^t$ to the language space as it does for a single image, yielding the intermediate compressed video token sequences $\mathbf{X}_V = [\mathbf{x}_V^1, \cdots, \mathbf{x}_V^t, \cdots, \mathbf{x}_V^T] \in \mathbb{R}^{T \times N \times C}$. As shown in Figure~\ref{fig:framework}, both the visual encoder and vision-language adapter are shared between image and video inputs.

Practically, we could directly input the video token sequences $\mathbf{X}_V$ to the LLM decoder. However, considering that videos are expanded in the temporal dimension compared with images, the number of tokens entering LLM will be much more than that of images, which will increase the computational overhead of the model. Meanwhile, since the content between video frames is continuous, there may be some redundancy in video tokens across different frames. Based on the above consideration, we propose to reduce the video token numbers by introducing a weighted token sampler module.

\begin{table}[tp]
\centering
\caption{A summary of image training datasets.}
\label{tab:image_training_data}
\resizebox{1.0\columnwidth}{!}
{
\begin{tabular}{l|l}
    \toprule
    Training Stage & Datasets  \cr
    \midrule
    multimodal aligning  & LAION~\cite{laion-400m}\cr
    \hline
    \multirow{3}{*}{multimodal pretraining}  
      &  GRIT~\cite{Kosmos2}, Visual Genome~\cite{krishna2017visual} \cr
      & VQAv2~\cite{goyal2017making}, GQA~\cite{hudson2019gqa}\cr
      &  LAION~\cite{laion-400m}, ShareGPT4V-caption~\cite{chen2023sharegpt4v} \cr
    \hline
    instruction tuning  & ShareGPT4V~\cite{chen2023sharegpt4v}, M3IT~\cite{li2023m3it} \cr
    \bottomrule
\end{tabular}
}
\end{table}

\begin{table}[tp]
\centering
\caption{The statistics of our video training datasets.}
\label{tab:video_training_data}
\resizebox{1.0\columnwidth}{!}
{
\begin{tabular}{cc|c|c}
    \toprule
    Dataset & Stage & Unique Video \textit{Num.} & QA pairs \textit{Num.}  \cr
    \midrule
    Valley702k~\cite{luo2023valley} & \multirow{2}{*}{pretraining} & 228,914 & 7,029,710 \cr
    -10\% &  & 22,891 & 86,314 \cr
    \hline
    VideoInstruct~\cite{maaz2023video} & \multirow{4}{*}{instruction} & 13,040 & 98,138 \cr
    -10\% & & 1,304 & 9,827 \cr
    -30\% &  & 3,912 & 29,396 \cr
    -60\% &  &7,824 & 58,950 \cr
    \bottomrule
\end{tabular}
}
\end{table}

\subsubsection{The Weighted Token Sampler Module}

Considering the video token feature $\mathbf{\hat{v}}^t \in \mathbb{R}^{M \times D}$ of the $t$-th video frame and one single learnable query token vector $\mathbf{q}^i \in \mathbb{R}^{1 \times C}$, after the cross-attention module, we could obtain the attention outputs of them:
\begin{equation}
    \mathbf{a}^t_i, \mathbf{x}_V^{t,i} = \mathcal{CA}(\mathbf{q}^i,\mathbf{\hat{v}}^t).
\end{equation}
Here $\mathbf{a}^t_i \in \mathbb{R}^{1 \times M}$ is the output attention weights, and $\mathbf{x}_V^{t,i} \in \mathbb{R}^{1 \times C}$ is the output attentive feature, which could be seen as the $i$-th token in $\mathbf{x}_V^t$.  Actually, $\mathbf{a}^t_i$ reflects the visual attention responses of $\mathbf{q}^i$ among the $M$ image patches in the $t$-th video frame. If the maximum value of $\mathbf{a}^t_i$ is still very low, it indicates that the query $\mathbf{q}^i$ does not obtain valid information from this video frame, so the value of the corresponding $\mathbf{x}_V^{t,i}$ will become relatively low and can be discarded in this case.  According to this idea, we can reduce the number of tokens in $\mathbf{x}_V^t$ as follows:
\begin{equation}
\begin{split}
\mathbf{r}^t = [max(\mathbf{a}^t_1),& \cdots, max(\mathbf{a}^t_i),  \cdots, max(\mathbf{a}^t_N)], \\
idx &= {\rm Argsort}(\mathbf{r}^t)[::-1][:K], \\
\mathbf{s}_V^t &= \mathbf{x}_V^t[idx] \in \mathbb{R}^{K \times C}. \\
\end{split}
\end{equation}
Based on the above equations, by only preserving those tokens that have top-$K$ higher visual responses, we reduce the tokens of the $t$-th video frame from $\mathbf{x}_V^t \in \mathbb{R}^{N \times C}$ to $\mathbf{s}_V^t \in \mathbb{R}^{K \times C}$ ($K$\textless $N$). Thus, we could correspondingly get the overall sampled video token feature sequences $\mathbf{S}_V = [\mathbf{s}_V^1, \cdots, \mathbf{s}_V^t, \cdots \mathbf{s}_V^T] \in \mathbb{R}^{T \times K \times C}$, which will be feed into the LLM decoder in the following. The overall model architecture can be found in Figure~\ref{fig:framework}.

\subsubsection{Video-Incorporated Training}

Besides the model architecture, the training strategy is also proven to be crucial for LVLMs~\cite{zhu2023minigpt,li2023llama,lin2023video,liu2023improved,luo2023valley}. Considering that the basic image-LVLM we constructed should lay a good foundation in visual semantic understanding, we consider using a small amount of video data during the video-LVLM training phase.  Our goal is to devise cost-effective training strategies that enhance performance. We categorize our training approaches into three distinct types:


\noindent \textbf{S4-V}: We directly add a fourth video instruction tuning stage to the image-LVLM after the three-stage training, and the trained modules in this process are consistent with the instruction tuning stage of image-LVLM. In this stage, we only use video data for training, and we denote it as the stage-4 tuning with video data only (S4-V).

\noindent \textbf{S3-IV}: In the above S4-V strategy, video data is introduced late in the process of model training, so we consider introducing a part of video data to guide the model early in the training process of image-LVLM, in order to obtain better video processing ability. Therefore, in the third instruction tuning stage of image-LVLM, we add a small amount of video instruction data, and jointly tune the model with the image instruction data. We denote such strategy as S3-IV.

\noindent \textbf{S2-S3-IV}: Furthermore, in order to explore the impact of introducing video data at an earlier pretraining stage on the model, we also design a joint pretraining and instruction tuning strategy. That is, in the second pretraining and the third instruction tuning stages of image-LVLM, a part of video pretrain/instruction data are also introduced for joint training, correspondingly. This strategy is named as S2-S3-IV.

For the video training data, we follow previous works and adopt the Valley702K dataset~\cite{luo2023valley} for pretraining and VideoInstruct~\cite{maaz2023video} for instruction tuning. The statistics of these datasets are provided in Table~\ref{tab:video_training_data}. In order to explore the influence of the amount of video data, we sample these datasets, and take 10\%, 30\%, 60\%, and 100\% videos and their corresponding QA-pairs for training, respectively. The detail results are shown in our experimental analysis.




\begin{table}[!t]
\center
\begin{small}
\resizebox{1.0\columnwidth}{!}
{
\begin{tabular}{ccc|cccc}
\toprule
\multirow{2}{*}{\textbf{Method}} & \multirow{2}{*}{\textbf{LLM}} & \multirow{2}{*}{\textbf{Res.}}   & \multicolumn{4}{c}{\textbf{Image Benchmark Toolkit}} \\
 & & & Ehub & MMB & MMB$^{\text{CN}}$& SEED$^\text{I}$ \\
 \midrule
mPLUG-Owl~\cite{ye2023mplug} & LLaMA-7B & 224 & 209.4 & 46.6 & - & - \\ 
Otter~\cite{li2023otter} & LLaMA-7B & 224 & - & 32.6 & - & - \\ 
IDEFICS-9B~\cite{idefics} &  LLaMA-7B & 224 & - & 48.2 & 25.2 & - \\ 
InstructBLIP~\cite{instructblip} & Vicuna-7B & 224 & 300.6 & 36.0 & 23.7 & 58.8 \\
Qwen-VL~\cite{bai2023qwen} & Qwen-7B & 448 & - & 38.2 & 7.4 & 62.3 \\
Qwen-VL-Chat~\cite{bai2023qwen} & Qwen-7B & 448 & 316.8 & 60.6 & 56.7 & 65.4 \\
MiniGPT-v2~\cite{chen2023minigpt} & LLaMA2 7B &448 & - & - & - & - \\ 
MiniGPT-v2-Chat~\cite{chen2023minigpt} & LLaMA2-Chat 7B & 448 & - & - & - & - \\ 
LLaVA-1.5~\cite{liu2023improved}& Vicuna-7B& 336 &  307.1 & 64.3 & 58.3 & - \\
InternLM-XComposer-VL~\cite{zhang2023internlm} & InternLM-7B & 224 & 322.5 & 74.8 & \textbf{73.1} & 66.9 \\ 
Video-LLaVA~\cite{lin2023video} & Vicuna-7B & 224 & - & 60.9 & - & - \\
LLaMA-VID~\cite{li2023llama} & Vicuna-7B & 336 & - & 65.1 & - & 67.6 \\ 
\midrule
FTFV-LLM$_{\text{image}}$ & Vicuna-7B & 448 & \textbf{325.2} & \textbf{75.5} & 68.1 & 69.5 \\
FTFV-LLM$_{\text{S4-V}}$ & Vicuna-7B & 448 & 318.2 & 74.6 & 68.8 & 69.0 \\
FTFV-LLM$_{\text{S3-IV}}$ & Vicuna-7B & 448 & \underline{323.8} & \textbf{75.5} & \underline{69.5} & \textbf{69.9} \\
FTFV-LLM$_{\text{S2-S3-IV}}$ & Vicuna-7B & 448 & 318.5 & \underline{75.2} & 68.5 & \underline{69.8} \\
\bottomrule
\end{tabular}
}
\end{small}
\caption{Comparison with SoTA methods on 4 image benchmarks. Res indicates input image resolution. Benchmark names are abbreviated due to space limits. Ehub: LVLM-Ehub~(Tiny)~\cite{shao2023tiny}; MMB: MMBench~(dev)~\cite{liu2023mmbench}; MMB$^{\text{CN}}$: MMBench-Chinese~(dev)~\cite{liu2023mmbench}; SEED$^\text{I}$: SEED-Bench~(image)~\cite{li2023seed}. The highest result is in \textbf{bold}, and the second highest is \underline{underlined}.}\label{tab:image_benchmark}
\end{table}

\begin{table*}[ht]
\center
\begin{small}
\resizebox{0.95\textwidth}{!}
{
\begin{tabular}{cccc|cccccc|cccc|c}
\toprule
\multirow{2}{*}{\textbf{Method}} & 
\multicolumn{3}{c|}{\textbf{Video Training Data}} & 
\multicolumn{2}{c}{\textbf{MSVD-QA}} & 
\multicolumn{2}{c}{\textbf{MSRVTT-QA}} & 
\multicolumn{2}{c|}{\textbf{ActivityNet-QA}} &
\multicolumn{4}{c|}{\textbf{Video-Bench}} &
\textbf{MVBench} \\
&Align& Pretrain& Instruct& Acc & Score & Acc & Score & Acc & Score & Exc & Prior & Dec & All & Avg \\
\midrule
FrozenBiLM~\cite{yang2022zero}&0 & 10,000k & 0 & 33.8 & - & 16.7 & - & 25.9 & - & - & - & - & - &-\\
VideoLLaMA~\cite{zhang2023video}&0 &2,500k & 11k& 51.6 & 2.5 & 29.6 & 1.8 & 12.4 & 1.1 & 32.5 & 27.8 & 38.2 & 32.8 & 34.1 \\
LLaMA-Adapter~\cite{zhang2023llama}&0 &0 & 0 & 54.9 & 3.1 &  43.8 & 2.7 & 34.2 & 2.7 & - & - & - & - & - \\
VideoChat~\cite{li2023videochat}&0 &10,000k & 11k & 56.3 & 2.8 & 45.0 & 2.5 & 26.5 & 2.2 & 34.1 & 29.6 & 42.5 & 35.4 & 35.5 \\
Video-ChatGPT~\cite{maaz2023video}&0 &0 & 100k& 64.9 & 3.3 & 49.3 &  2.8 & 35.2 & 2.7 & 39.8 & 29.2 & 46.5 & 38.5 & 32.7 \\
BT-Adapter~\cite{liu2023one}&0 &2,000k & 100k & 67.5 & 3.7 & 57.0 & 3.2 & 45.7 & 3.2 & - & - & - & - & -  \\
Otter~\cite{li2023otter}&0 &0 &502k & - & - & - & - & - & - & 37.5 & 33.0 & 41.9 & 37.5 & 26.8 \\
PandaGPT~\cite{su2023pandagpt}& 0&0 &0 & - & - & - & - & - & - & 37.5 & 32.0 & 43.1 & 37.5 & - \\
mPLUG-Owl~\cite{ye2023mplug}& 0&0 &0 & - & - & - & - & - & - & 33.2 & 26.4 & 39.9 & 33.2 & 29.7 \\
Video-LLaVA~\cite{lin2023video}&10,000k &702k &100k  & \textbf{70.7} & \textbf{3.9} & \textbf{59.2} & \textbf{3.5} & 45.3 &  \underline{3.3} & - & - & - & - & - \\
LLaMA-VID~\cite{li2023llama}&0 &232k & 98k & 69.7 & 3.7 & \underline{57.7} & 3.2 & 47.4 & \underline{3.3} & - & - & - & - & - \\
VideoChat2~\cite{li2023mvbench}&10,000k &10,000k &800k  & 70.0 & \textbf{3.9} & 54.1 & 3.3 & 49.1 & \underline{3.3} & - & - & - & - & \textbf{51.1} \\
\midrule
FTFV-LLM$_{\text{image}}$ (ours)&0 &0 & 0 & 65.0 &3.6  &42.9  &2.6  &46.9  &3.2 & 52.4 & \textbf{42.8} & 52.3 & 49.2 & 42.4 \\
FTFV-LLM$_{\text{S4-V}}$ (ours)&0 &0 &9.8k & 69.2 & 3.7 & 55.1 & 3.1 & 49.7 & \underline{3.3} & \textbf{54.3} & 42.0 & \textbf{54.5} & \textbf{50.3} & 43.5 \\
FTFV-LLM$_{\text{S3-IV}}$ (ours) &0&0 &9.8k  & \underline{70.3} &  \underline{3.8} & 53.4 & 3.1 & \textbf{52.2}  & \textbf{3.4} & \underline{53.2} & 41.9 & \underline{53.9} & 49.7 & \underline{44.8} \\
FTFV-LLM$_{\text{S2-S3-IV}}$ (ours)&0 &86.3k & 9.8k & 68.9 & 3.7 & 52.0 & 3.1 & \underline{52.0} & \textbf{3.4} & \textbf{54.3} & \underline{42.5} & 53.6 & \underline{50.1} & 44.2 \\
\bottomrule
\end{tabular}
}
\end{small}
\caption{Comparison with leading methods on three Video-QA benchmarks, Video-Bench~\cite{ning2023video} and MVBench~\cite{li2023mvbench}. Notably, in each training strategy (S4-V, S3-IV, S2-S3-IV) of our FTFV-LLM, we only use a small amount of video data compared to other methods~(about 10\% of ~\cite{lin2023video,li2023llama}), with 128 tokens per video frame. Subset names in Video-Bench are abbreviated due to page limitations. Exc: Video-Exclusive Understand; Prior: Prior Knowledge-based Question-Answering; Dec: Comprehension and Decision-Making.  The highest result is in \textbf{bold}, and the second highest is \underline{underlined}.} 
\label{tab:video_benchmark}
\end{table*}

\begin{table*}
\center
\begin{small}
\resizebox{0.85\textwidth}{!}
{
\begin{tabular}{cc|cccccc|c|c}
\toprule
\multirow{2}{*}{\textbf{Method}} & 
\multirow{2}{*}{\textbf{TFLOPs}} & 
\multicolumn{2}{c}{\textbf{MSVD-QA}} & 
\multicolumn{2}{c}{\textbf{MSRVTT-QA}} & 
\multicolumn{2}{c|}{\textbf{ActivityNet-QA}} & 
\multicolumn{1}{c|}{\textbf{Video-Bench}} & 
\textbf{MVBench}\\
&& Acc & Score & Acc & Score & Acc & Score &All & Avg \\
\midrule

FTFV-LLM$_{\text{S4}\text{-V}_{4}}$ &32.14  & 67.31 & 3.69 & 49.61 &3.00  & 44.85 &  3.16&  49.65& 44.57 \\
FTFV-LLM$_{\text{S4}\text{-V}_{16}}$ &33.47  & 68.29 & 3.74 & 53.45 &3.16  & 48.44 & 3.31 & 49.59 & 44.70 \\
FTFV-LLM$_{\text{S4}\text{-V}_{32}}$ & 35.24 & 68.32 &  3.73 & 54.07 & 3.17 & 49.43 & 3.32 & 49.87&44.38 \\
FTFV-LLM$_{\text{S4}\text{-V}_{64}}$ &38.79   & 68.85 & 3.74 & 54.80 & 3.19 &  49.46& 3.32 & 49.84&44.07\\
FTFV-LLM$_{\text{S4}\text{-V}_{128}}$ &45.88 & 69.18 & 3.73 & 55.06 & 3.14 & 49.67 & 3.31 & 50.26& 43.52 \\
FTFV-LLM$_{\text{S4}\text{- V}_{256}}$ &60.07 & 68.62 & 3.71& 53.28 &3.13  & 48.17 & 3.30&49.39 &44.47 \\

\bottomrule
\end{tabular}
}
\end{small}
\caption{Exploring the impact of sampled token numbers in the weighted token sampler of FTFV-LLM. FTFV-LLM$_{\text{S4}\text{-V}_{x}}$ means we sample $x$ tokens in the weighted token sampler module, and the model is trained with the S4-V training strategy. FLOPs stands for the number of floating-point operations.}
\label{tab:impact_of_weighted_token_sampler}
\end{table*}



\section{Experiments}
\subsection{Experimental Setup}

\noindent \textbf{Implementation Details.} In this work, we instantiate the ViT visual encoder and the vision-language adapter with the pretrained weights of QWen-VL~\cite{bai2023qwen}, and the LLM decoder with Vicuna-7B~\cite{vicuna2023}. In accordance with the previous section~\ref{capter: 3.1.2}, we implemented a three-stage training strategy of image-based LVLMs and explored an efficient training approach for video-based LVLMs by incorporating varying proportions of video training data in each stage. 
In the first stage, the model underwent training on 64$\times$NVIDIA A100 GPUs with a learning rate (lr) of 2e-4. Subsequently, in the second and third stages, models were trained using 16$\times$NVIDIA A100 GPUs, with respective lr values of 5e-5 and 2e-5, respectively. Furthermore, for the S4-V training strategy, a dedicated stage was introduced to facilitate rapid video fine-tuning, employing 2$\times$NVIDIA A100 GPUs with a learning rate of 2e-5. We train 1 epoch in each stage and utilize LoRA\cite{hu2021lora} to efficiently train the model. For each video input, we uniformly sample 8 frames for training and testing. 

\subsection{Results on Image Benchmarks}

We first evaluate the basic image-LVLM (FTFV-LLM$_{\text{image}}$) of our model for image understanding on four image benchmark toolkits, the results are shown in Table~\ref{tab:image_benchmark}. These benchmark toolkits provide a detailed assessment of the model’s capabilities
through robust evaluation metrics. Meanwhile, we also report our FTFV-LLM with video-incorporated training in different training stages, \textit{i.e.,} FTFV-LLM$_{\text{S4-V}}$, FTFV-LLM$_{\text{S3-IV}}$, FTFV-LLM$_{\text{S2-S3-IV}}$. In these settings, we only use 10\% video data at each stage as reported in Table~\ref{tab:video_training_data}. We can observe that our FTFV-LLM$_{\text{image}}$ achieves the highest results on both Ehub and MMBench. And on SEED-Bench, FTFV-LLM$_{\text{image}}$ also outperforms other image-LVLMs. On MMBench-Chinese, the FTFV-LLM$_{\text{image}}$ is only inferior to InternLM-XComposer-VL, probably because we do not have richer Chinese training data as they do. These observations indicate that our FTFV-LLM$_{\text{image}}$ possesses an advanced comprehension of the semantic elements within visual scenes, which allows it to effectively address a wide range of open-ended and free-form natural language questions about images, and it also provides a good basis for video semantic understanding.

As for the video-LVLMs, on the Ehub benchmark, FTFV-LLM$_{\text{S4-V}}$, FTFV-LLM$_{\text{S3-IV}}$, FTFV-LLM$_{\text{S2-S3-IV}}$ lose a little bit of performance compared to FTFV-LLM$_{\text{image}}$, but on MMBench, MMBench-Chinese, and SEED-Bench, these video-LVLMs achieves competing results and even superior results than FTFV-LLM$_{\text{image}}$. This demonstrates that the integration of video data in our training strategy does not significantly diminish the model's image understanding capabilities.

\begin{figure*}
\centering
\includegraphics[scale=0.52]{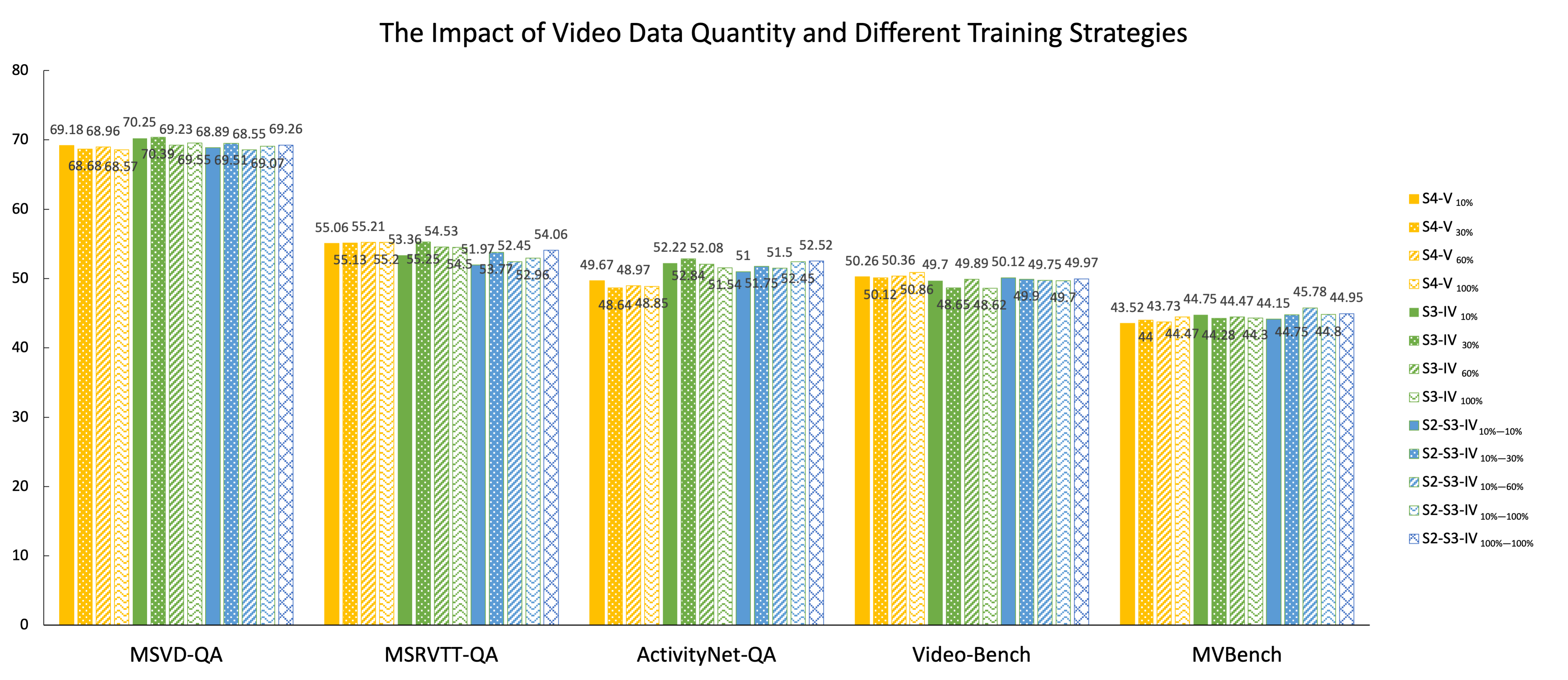}
\caption{Exploring the effects of varying video data proportions and training strategies. Here, S4-V$_{x\%}$ means we use $x$\% video data in the S4-V video instruction tuning stage, S3-IV$_{x\%}$ indicates we use $x$\% video data in the S3-IV stage, and the S2-S3-IV$_{x\%-y\%}$ refers that we incorporate $x$\% pretraining and $y$\% instruction tuning video data in the second and third model training stages, respectively. All stages incorporate the complete set of image data as shown in Table~\ref{tab:image_training_data}.}
\label{fig:impact_of_videodata_amount}
\end{figure*}

\subsection{Results on Video Benchmarks}

In this section, we evaluate FTFV-LLM on multiple video benchmarks. We first conduct evaluations on three common video QA benchmarks: MSVD-QA~\cite{chen2011collecting}, MSRVTT-QA~\cite{xu2016msr}, and ActivityNet-QA~\cite{caba2015activitynet}.
The three video-QA datasets all depict scenes from human daily life, encompassing many scenarios that require visual perception and action recognition. Additionally, we also perform evaluations on two newly proposed video benchmarks. These include a video benchmark named MVBench~\cite{li2023mvbench} which focuses on evaluating temporal understanding in dynamic video tasks, and a comprehensive video understanding benchmark named Video-Bench~\cite{ning2023video} which evaluates video-LVLMs across three distinct levels: Video-exclusive Understanding, Prior Knowledge-based Question-Answering, and Comprehension and Decision-making.

\noindent \textbf{Compared to Other Video-LVLMs.} Table \ref{tab:video_benchmark} compares our FTFV-LLM with leading video-LVLMs. Additionally, we assess our image-based model (FTFV-LLM$_{\text{image}}$) on these benchmarks by randomly sample a single frame from each video as input. FTFV-LLM$_{\text{image}}$ underperforms on the video-QA datasets and MVBench but excels on Video-Bench, particularly in subsets requiring prior knowledge. This is expected as FTFV-LLM$_{\text{image}}$, untrained on video data, lacks action recognition and temporal reasoning skills. Yet, its rich visual knowledge from image data provides an advantage on Video-Bench, suggesting the benchmark's lesser emphasis on video-specific queries.

For the video-LVLM of our model FTFV-LLM under each training strategy (S4-V, S3-IV, S2-S3-IV), we only use 10\% of the video training data as stated in Table~\ref{tab:image_training_data}, which is largely less than other methods~\cite{lin2023video,luo2023valley,li2023llama}. Meanwhile, we only use 128 tokens to represent each video frame. This is the default setting for our model, if not otherwise stated. Overall, we could observe that FTFV-LLM demonstrates exceptional performance across all datasets, although we only incorporate a few video training data to enhance the video understanding ability over the basic image-LVLM. Specifically, on the ActivityNet-QA dataset, the FTFV-LLM$_{\text{S3-IV}}$ achieves the highest results, surpassing the previous leading approach ~\cite{li2023mvbench} by a gain of 3.1\%, even though it is trained with 80 times the amount of video data compared to ours. Meanwhile, on the MSVD-QA dataset, the FTFV-LLM$_{\text{S3-IV}}$ achieves the second-best results. However, on the MSRVTT-QA our FTFV-LLMs do not perform as well as on the previous two benchmarks, but it is still in the top three and not far from the best. It is worth mentioning that the video-QA benchmarks' scores were determined by Chat-GPT, introducing some scoring variability.

Different from the QA benchmark, both Video-Bench and MVBench are organized in the form of multiple choice questions, and the model accuracy is computed based on comparing the model output options and ground truth options. This evaluation is relatively more straightforward than GPT scoring. As shown in Table~\ref{tab:video_benchmark}, we could observe that our FTFV-LLM$_{\text{S4-V}}$ achieves the highest score on Video-Bench, significantly surpassing the previous best approach Video-ChatGPT by 11.8\%, demonstrate the effectiveness of our FTFV-LLM in processing more complex and comprehensive scenes.


On MVBench, which emphasizes temporal comprehension in dynamic videos, our FTFV-LLM secures second place, significantly outperforming VideoChat (by 9.3\%) and VideoLLaMA (by 10.7\%), showcasing our model's capability in grasping video temporality. Remarkably, FTFV-LLM achieves this without any dedicated temporal modeling modules, relying solely on the LLM's inherent context and sequence processing strengths to interpret temporal relationships in videos. This impressive result suggests that with a solid foundation in image understanding and the LLM's sequential processing, there is substantial potential to develop a competent video-LVLM for modeling video temporal dynamics. However, there remains a gap between FTFV-LLM and the top-performing method on MVBench, VideoChat2, which may be attributed to VideoChat2's more advanced video encoder UMT-L~\cite{li2023unmasked} and its use of more extensive video training data. This indicates that while FTFV-LLM can yield promising outcomes on general video understanding tasks with just a strong image-LVLM and limited video training. As for more intricate temporal tasks, integrating superior video encoding techniques with LLM and utilizing a broader range of video data would lead to enhanced performance.

\subsection{Ablation Studies}

\noindent \textbf{The Impact of Video Data Quantity.}
Figure \ref{fig:impact_of_videodata_amount} illustrates the performance of various training strategies (S4-V, S3-IV, S2-S3-IV) on video benchmarks with different video data volumes. For S4-V, the performance variation is minimal across benchmarks when altering the amount of video data used in training. On the MSVD-QA benchmark, the largest observed difference is only 0.61\%, while on MSRVTT-QA, ActivityNet-QA, Video-bench, and MVBench, it is just 0.15\%, 1.03\%, 0.74\%, and 0.95\%, respectively. Notably, the S4-V$_{10\%}$ model, trained with merely 10\% of the video data, still outperforms others on MSVD-QA and ActivityNet-QA. This pattern holds for S3-IV and S2-S3-IV strategies as well, where S2-S3-IV$_{10\%-10\%}$ denotes using 10\% of video data for both pretraining and instruction phases, compared to S2-S3-IV$_{100\%-100\%}$ with full video data usage, showing a negligible performance gap. Overall, the influence of video data quantity on training outcomes is slight for all strategies, likely due to the extensive image data incorporated in training, providing a solid base for semantic visual understanding. Consequently, only a small amount of video data is necessary to bolster the model's video comprehension capabilities.


\noindent \textbf{The Impact of Different Training Strategies.}
Figure~\ref{fig:impact_of_videodata_amount} also allows us to assess our FTFV-LLM across various training strategies. The yellow bars denote the S4-V strategy, green bars for S3-IV, and blue for S2-S3-IV. These strategies differ primarily in the timing of video data integration during training. S4-V introduces video data last, adding a video-focused instruction tuning stage after completing basic image-LVLM training. S3-IV incorporates video and image data simultaneously during the third instruction tuning stage, while S2-S3-IV starts the earliest, blending video with image data during both the second pretraining and third instruction tuning stages. S4-V offers the simplest extension from image-LVLM to video-LVLM, requiring only an additional training phase without altering the image-LVLM training. In contrast, the other strategies involve some retraining of the image-LVLM. According to Figure~\ref{fig:impact_of_videodata_amount}, the performance discrepancies among the three strategies on video benchmarks are marginal. On the MSVD-QA and ActivityNet-QA datasets, S3-IV and S2-S3-IV slightly outperform S4-V, whereas S4-V leads on the MSRVTT-QA and Video-Bench.


The observed patterns are intriguing. It is commonly assumed that introducing video data early on should bolster a model's video understanding and thus enhance video-LVLM performance. However, our results do not align with this expectation. We believe this could be due to two main factors. First, the current video training and evaluation datasets tend to emphasize general visual comprehension (such as scenes, objects, spatial relations, etc.) and often overlook the assessment of temporal dynamics in videos~\cite{li2023mvbench}. As a result, the evaluation of video-LVLMs does not differ significantly from that of image-LVLMs, rendering the timing of video data introduction relatively inconsequential under the present evaluation metrics. Second, given that video training data generally falls short of image data in both volume and diversity, incorporating video data too early might actually dilute the model's generalizability. This is evident from Table \ref{tab:image_benchmark}, where FTFV-LLM$_{\text{S2-S3-IV}}$ shows a drop in performance compared to FTFV-LLM$_{\text{image}}$. Hence, in this context, more or earlier exposure to video data does not necessarily translate to a benefit.


\noindent \textbf{The Impact of Weighted Token Sampler.} Table \ref{tab:impact_of_weighted_token_sampler} presents the effects of employing various token numbers in our weighted token sampler on different video benchmarks. We can observe that the disparity in performance between the 16-token variant (FTFV-LLM$_{\text{S4-V}_{16}}$) and the 256-token variant (FTFV-LLM$_{\text{S4-V}_{256}}$) is marginal across all the benchmarks, with the largest discrepancy being approximately 1.8\%. A reduction to 4 tokens results in a more pronounced performance difference, particularly noticeable in the video-QA dataset. These findings indicate that selecting tokens based on higher attention weights can maintain performance while substantially cutting computational costs. For instance, model computation complexity drops from 60.07 TFLOPs to 33.47 TFLOPs when token count per frame is reduced from 256 to 16, nearly halving the computation. This improvement in efficiency not only conserves resources but also enhances the model's capability to process longer video sequences with greater efficacy.



\begin{table}
\center
\begin{small}
\resizebox{0.46\textwidth}{!}
{
\begin{tabular}{ccc|cc|c|c}
\toprule
\multirow{2}{*}{\textbf{Method}} & 
\multirow{2}{*}{\textbf{Data}} &
\multirow{2}{*}{\textbf{Type}} & 
\multicolumn{2}{c|}{\textbf{ActivityNet-QA}} & 
\multicolumn{1}{c|}{\textbf{Video-Bench}} & 
\textbf{MVBench} \\
& &  &Acc &Score  & All &Avg\\
\midrule
VideoChat2~\cite{li2023mvbench} &800k & \multirow{2}{*}{\textit{all}}  &\underline{49.1} &\underline{3.3} &- &\textbf{51.1}  \\
FTFV-LLM$_{\text{S4}\text{-V}}$  & 10k &  & 48.0 &\underline{3.3} &\underline{50.2} & 45.1 \\
\midrule
FTFV-LLM$_{\text{S4}\text{-V}}$  &10k & \textit{classification} & 17.3 & 1.8 & 48.2 &45.4  \\
FTFV-LLM$_{\text{S4}\text{-V}}$ &10k &\textit{simple caption}   & 42.1 & 2.9 & 49.4 &45.1  \\
FTFV-LLM$_{\text{S4}\text{-V}}$  &10k & \textit{detailed caption}   &49.0  &\underline{3.3}  &49.7 &43.6 \\
FTFV-LLM$_{\text{S4}\text{-V}}$ &10k & \textit{conversation}    &47.7  &3.2  &\textbf{50.3} & 43.3 \\
FTFV-LLM$_{\text{S4}\text{-V}}$  & 10k  
 &  \textit{vqa} &\textbf{51.1} &\textbf{3.4} & 49.7&43.3 \\
FTFV-LLM$_{\text{S4}\text{-V}}$  &10k & \textit{reasoning}  &40.5 &3.0 &49.4 & \underline{46.0} \\
FTFV-LLM$_{\text{S4}\text{-V}}$  &10k & \textit{vqa+resoning}  &47.3 &\underline{3.3} &49.9 & 45.6 \\
\bottomrule
\end{tabular}
}
\end{small}
\caption{
Examine the impact of video instruction tuning data types on model efficacy. `Data' refers to the quantity of video QA pairs utilized during the instruction tuning phase. `Type' denotes the category of video instruction data, encompassing six distinct types: \textit{classification}, \textit{simple caption}, \textit{detailed caption}, \textit{conversation}, \textit{vqa}, and \textit{reasoning}. The type \textit{all} indicates a mixed training approach using a combination of all six video data categories.  The highest result is in \textbf{bold}, and the second highest is \underline{underlined}.
}
\label{tab:video_instruction_data_type}
\end{table}

\begin{figure*}
\centering
\includegraphics[scale=0.6]{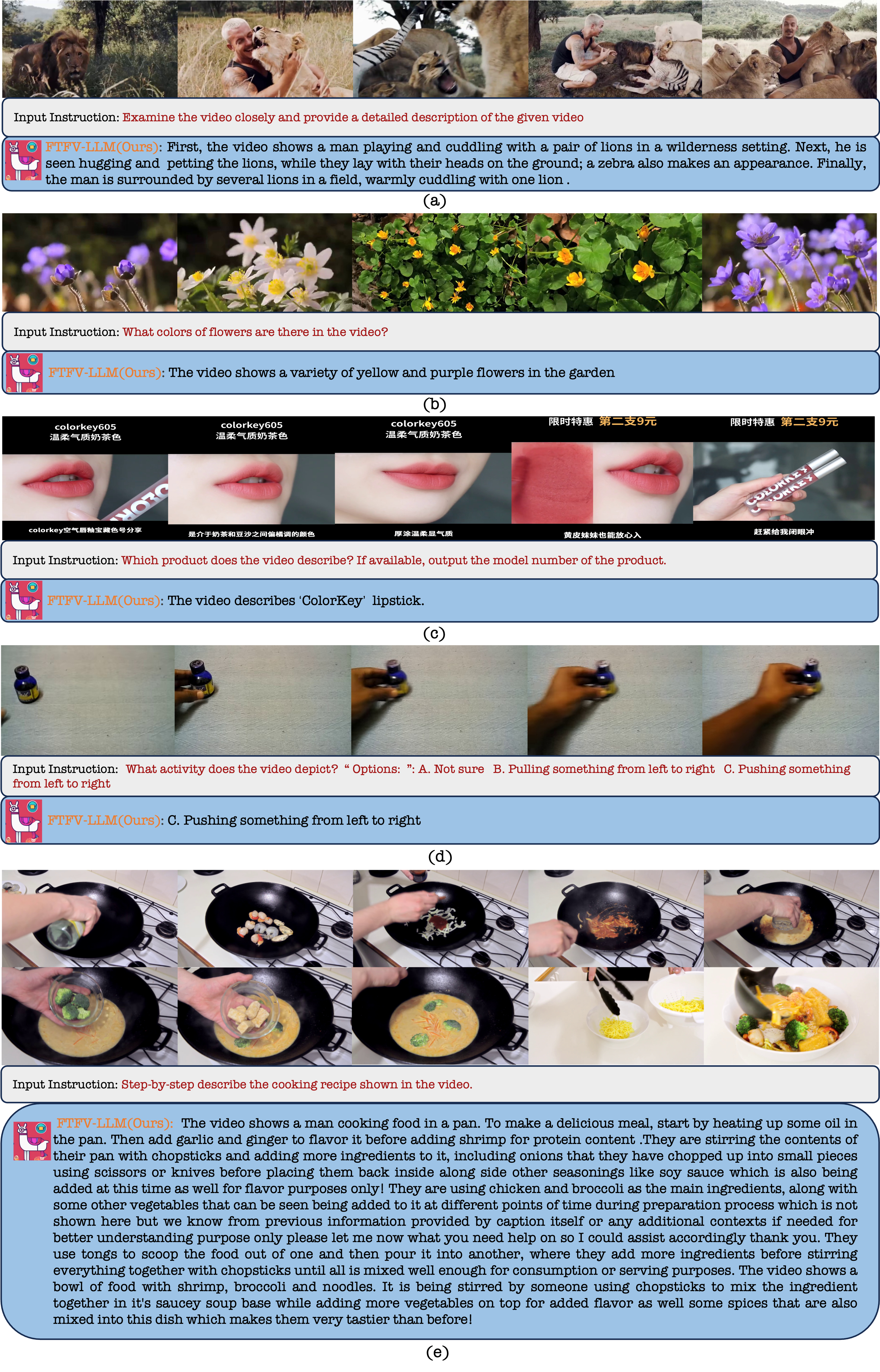}
\caption{Qualitative examples for our FTFV-LLM.}
\label{fig:qualitative_results}
\end{figure*}

\subsection{The Investigation of Video Instruction Data}

The recent VideoChat2 model~\cite{li2023mvbench} introduces a comprehensive video instruction tuning dataset, VideoChat2-IT, which includes a variety of video data categorized into six types: \textit{vqa}, \textit{reasoning}, \textit{detailed caption}, \textit{simple caption}, \textit{conversation}, and \textit{classification}. By leveraging this diverse range of video instruction data, VideoChat2 has demonstrated notable success. Inspired by VideoChat2-IT's meticulous categorization, we aim to identify which video instruction data types most effectively enhance the transition from an image-LVLM to a video-LVLM. Thus, we experiment with different data types from VideoChat2-IT during the S4-V stage of our base FTFV-LLM$_{\text{image}}$, while maintaining a consistent volume of 10k QA pairs for each data type. Additionally, we train our FTFV-LLM using a random mix of 10k video QA pairs across all six data categories. The findings of these experiments are detailed in Table~\ref{tab:video_instruction_data_type}.

Firstly, in the top part of Table~\ref{tab:video_instruction_data_type}, we note that our FTFV-LLM, trained with a mere 10k video QA pairs from all six data types, delivers performance on par with VideoChat2 on ActivityNet-QA and Video-Bench. This reaffirms that even a small quantity of video data can significantly enhance an image-LVLM's capabilities in the video domain. However, there remains a performance gap on MVBench when compared to VideoChat2, likely due to VideoChat2's utilization of a more advanced video encoder, which confers an edge in benchmarks like MVBench that emphasize temporal inference.

Secondly, in the bottom part of Table~\ref{tab:video_instruction_data_type}, we observe that different types of video data used for S4-V tuning impact the FTFV-LLM's performance on ActivityNet-QA and MVBench significantly, while results on Video-Bench remain relatively stable. Training with \textit{simple caption} and \textit{classification} leads to a marked decrease in ActivityNet-QA accuracy, likely due to their limited and basic instructional content. Training with \textit{detailed caption} and \textit{conversation} also fails to yield improvements on ActivityNet-QA and results in lower MVBench scores. These four data types lack explicit emphasis on video temporal reasoning, offering only rudimentary visual understanding akin to what might be gleaned from image datasets, thus their contribution to video-LVLM training is relatively minor.

Notably, by using exclusively \textit{vqa} data, our FTFV-LLM outperforms VideoChat2 on ActivityNet-QA, suggesting a strong fit for video question answering. However, this specialization comes at the cost of temporal understanding, as evidenced by the lowest score of 43.3\% on MVBench. In contrast, \textit{reasoning} data leads to lower scores on ActivityNet-QA but the best on MVBench, indicating that temporal reasoning training enhances the model's ability to understand video dynamics. However, the \textit{reasoning} data contains a large proportion (about 70\%) of synthetic data from CLEVR~\cite{johnson2017clevr} with limited and single scene, and the QA pairs of it are also mostly in the form of multiple choice questions instead of open-world QAs, the model trained on such data loses the generalization ability from the basic FTFV-LLM$_{\text{image}}$, and thus does not achieve good results on ActivityNet-QA.

As shown in the last row of Table ~\ref{tab:video_instruction_data_type}, merging \textit{vqa} and \textit{reasoning} data yields a model whose performance matches that trained on all six data types across benchmarks. This suggests that the varied visual scenes and open-ended diverse instructions in \textit{vqa} and the temporal reasoning aspects of \textit{reasoning} data are complementary. It also indicates that the variety of video training data needed when transitioning from an image-LVLM to a video-LVLM does not have to be extensive. The foundational visual knowledge often present in video data may already be covered during image training. Our findings highlight the significance of incorporating video data that offers temporal reasoning insights absent in image datasets. Additionally, the scene diversity in videos, the variety in QA pair instructions, and the range of question answering formats are essential for preserving the model's generalization ability.

\subsection{Qualitative Results}
As depicted in Figure ~\ref{qualitative_results}, we provide a series of case studies to evaluate our model's performance. Subfigure (a) demonstrates the model's precise identification of actions and interactions between humans and animals within the video, succinctly summarizing the primary narrative and showcasing the model's adeptness at generating accurate video captions. In subfigure (b), the model's ability to discern the color of flowers is presented, illustrating its capacity to recognize detailed object attributes with high fidelity. Further, subfigure (c) exhibits the FTFV-LLM's accurate identification of products and their respective model numbers, suggesting a promising application for the model in Optical Character Recognition (OCR) tasks. In subfigure (d), the model's proficiency in detecting motion direction is assessed, with the FTFV-LLM displaying commendable performance in capturing such movement.

In subfigure~(e), the model's understanding of action sequences in long videos is scrutinized. While the model adeptly recognizes sequences of actions over time, it is not immune to instances of hallucination, erroneously incorporating unrelated objects such as `chicken', `chopsticks', and `soy sauce'. This phenomenon is indicative of a prevalent challenge across contemporary Large Language Models (LLMs) and delineates a pivotal direction for enhancement in subsequent research endeavors.

\section{Conclusion}


In summary, our research extends a basic image-LVLM to a video-LVLM by utilizing the inherent visual connections between the two modalities. We have developed a cost-efficient video-LVLM with an optimized architecture, innovative training strategies, and targeted video instruction data investigation.  Our proposed weighted token sampler efficiently compresses the video tokens of each video frame, cutting computational costs while maintaining model performance. With just 10\% of the usual video training data, our model still still achieved impressive results across multiple training stages. The importance of video data for temporal reasoning is also underscored. Our Fewer Tokens and Fewer Videos Language-Vision Language Model (FTFV-LVLM) excels in both image and video benchmarks, validating our methodology and contributing to the progress in video comprehension using minimal data.



{
    \small
    \bibliographystyle{ieeenat_fullname}
    \bibliography{main}
}


\end{document}